\documentclass[journal]{IEEEtran}

\ifCLASSINFOpdf
\else
\fi
\usepackage{amsmath,amssymb}  

\usepackage{algorithm}
\usepackage{algpseudocode}
\usepackage{epsfig} 
\usepackage{amsmath} 
\usepackage{amssymb}  
\graphicspath{{../figs/}}
\usepackage{caption}
\usepackage{booktabs}
\usepackage{amsmath}
\usepackage{etoolbox}
\makeatletter
\patchcmd{\@makecaption}
  {\scshape}
  {}
  {}
  {}
\makeatother
\usepackage{orcidlink}
\usepackage{multirow}
\usepackage{subcaption}
\usepackage[table]{xcolor}
\begin{document}
%
\title{Graph-MambaNav: Spatial-Temporal Graph Mamba Leveraging Object-Relation Knowledge for Object-Goal Navigation}
%
%
%


\author{
Leyuan Sun$^{1,*}$\,\orcidlink{0000-0001-6123-9339},
Genxin Chen$^{1}$,
Linwei Ye$^{1}$,
Yan Zhang$^{1,2}$,
Xi Kan$^{1}$,
and Yanfei Sun$^{3,4}$%
\thanks{This work was supported by the Research Start-up Fund for High-level Talents of Wuxi University under Grant 2026r004 and by the National Natural Science Foundation of China under Grant XXXXXXXXXX.}%
\thanks{$^{1}$Leyuan Sun, Genxin Chen, Linwei Ye, Yan Zhang, and Xi Kan are with the School of Internet of Things Engineering, Wuxi University, Wuxi, Jiangsu 214105, China.}%
\thanks{$^{2}$Yan Zhang is also with the School of Communications and Information Engineering, Nanjing University of Posts and Telecommunications, Nanjing, Jiangsu 210003, China.}%
\thanks{$^{3}$Yanfei Sun is with the Wuxi Key Laboratory of Artificial Intelligence and Security, Wuxi, Jiangsu 214105, China.}%
\thanks{$^{4}$Yanfei Sun is also with Wuxi University, Wuxi, Jiangsu 214105, China.}%
\thanks{$^{*}$Corresponding author: Leyuan Sun (e-mail: sunleyuan@cwxu.edu.cn).}%
}
\maketitle

\begin{abstract}
Object-goal navigation requires an agent to reason over object relationships and prioritize target-relevant objects for efficient decision making in unseen environments. 
While existing graph-based methods incorporate target-awareness at the feature or attention level, they remain permutation-invariant and lack an explicit mechanism to control information propagation order, limiting their ability to model target-dependent importance and long-range dependencies. 
In contrast, Graph-Mamba highlights that node prioritization through sequence ordering is critical for effective global reasoning.
In this work, we investigate the node prioritization mechanism in Graph-Mamba and study its role in object navigation. 
We propose Graph-MambaNav, a target-aware spatial-temporal graph encoding framework that introduces a heuristic ordering over objects based on their relevance to the target, allowing more informative objects to be processed later to aggregate richer context. 
Both node ordering and edge weights are initialized from LLM-derived commonsense object relationships, providing a unified prior for structured reasoning. 
A spatial module integrates local message passing with global GraphMamba-based selective scanning, while a temporal module applies Mamba-based sequence modeling over object-wise temporal orders, allowing selective aggregation of historical context for long-range temporal reasoning.
Experiments on AI2-THOR and RoboTHOR demonstrate improved navigation performance with generalization, and additional real-world robot deployment further validates the effectiveness of our proposed approach.
\end{abstract}

\begin{IEEEkeywords}
Object-goal Navigation; Graph Sequence Modeling; Mamba; Node Prioritization; Commonsense Reasoning.
\end{IEEEkeywords}

%
\IEEEpeerreviewmaketitle

\section{Introduction}
\IEEEPARstart{I}{n} object-goal navigation (ObjectNav), agents must locate a target object in unseen environments using only egocentric observations. 
This requires not only recognizing objects but also reasoning about their spatial-temporal and semantic relationships to infer where the target is likely to be found. 
For example, to find a \textit{remote control}, an agent may first identify a \textit{television} and search its vicinity. 
Such relational reasoning naturally motivates the use of object-centric graphs  \cite{chi2025navigating} to model structured interactions among objects for navigation.

\begin{figure}[t]
\centering
\includegraphics[width=0.48\textwidth]{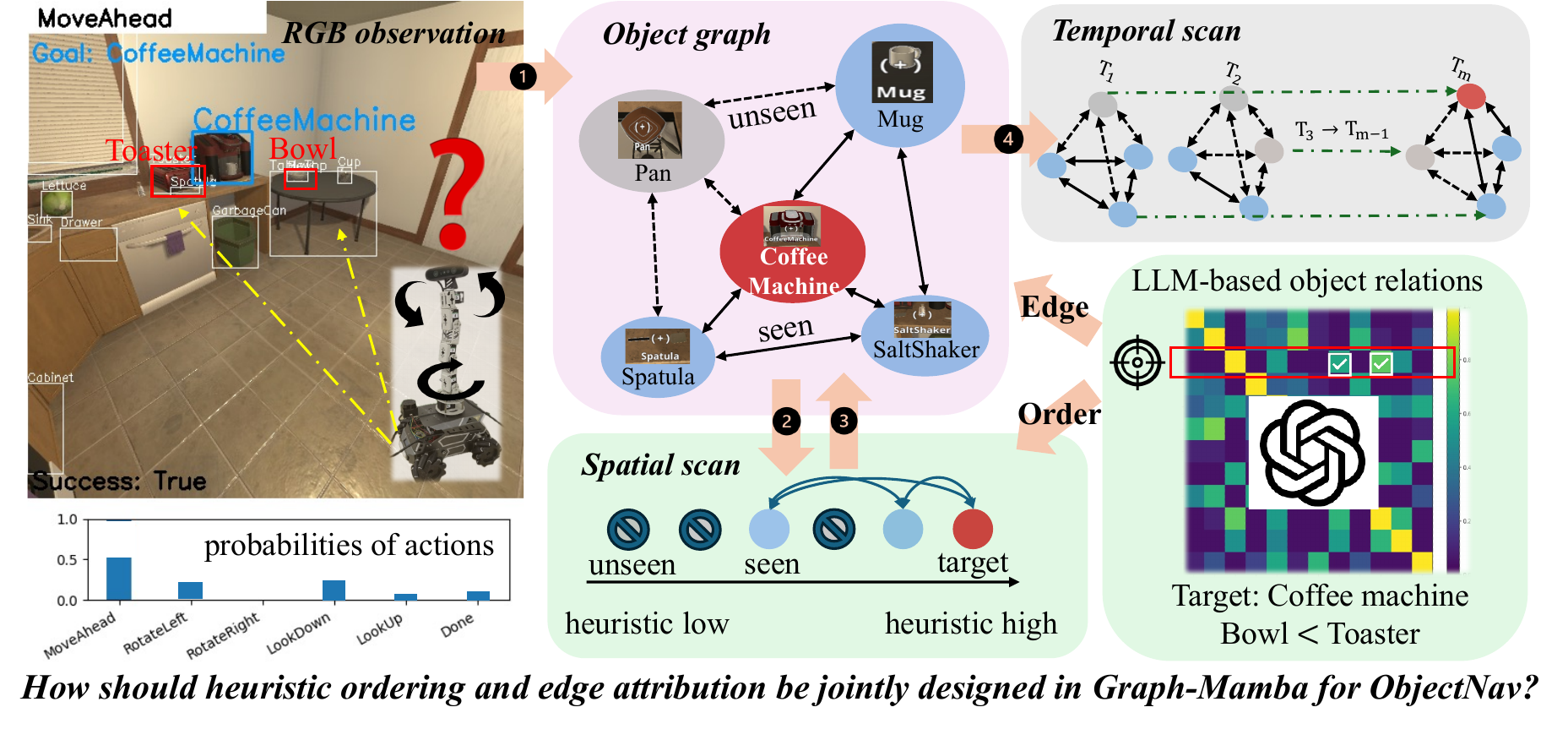}
\caption{
Motivation and overview of Graph-MambaNav. 
Given an egocentric RGB observation, an object-centric graph is constructed from detected objects. 
\textbf{LLM-derived commonsense relations} provide a unified prior for both edge attribution and \textbf{target-aware heuristic ordering}. 
The ordered spatial scan enables more target-relevant objects to aggregate richer context under Mamba-based scanning, while the temporal scan models object-wise dynamics across time for action prediction.
}
\label{intro}
\vspace{-0.6cm}
\end{figure}

Existing works construct object graphs and apply graph neural networks (GNNs) or graph attention mechanisms to learn such relationships \cite{zhang2021hierarchical}\cite{du2020learning}\cite{dang2022unbiased}. 
However, these approaches are inherently permutation-invariant and primarily rely on local message passing, lacking an explicit mechanism to control information propagation based on the target. 
Consequently, all objects contribute equally during initial computation, making it difficult to prioritize informative objects and capture long-range dependencies for goal-directed reasoning.

Recent advances in Graph-Mamba \cite{wang2024graph} introduce a new paradigm for graph modeling by leveraging State Space Models (SSMs) \cite{gu2021efficiently} with selective scanning, enabling efficient long-range dependency modeling. 
A key insight of Graph-Mamba is that node prioritization through sequence ordering allows more informative nodes to access richer contextual information during recurrent updates. 
However, this mechanism is designed for general graph tasks, where node importance is typically defined using heuristic criteria (e.g., node degree), and its role in task-driven scenarios such as object-goal navigation remains underexplored. 
In particular, beyond node ordering, defining edge weights that capture target-dependent object relationships is equally crucial for effective graph reasoning. 
As illustrated in Fig.~\ref{intro}, the above observations raise a fundamental question for this study: \textbf{\textit{How should heuristic ordering and edge attributions be jointly designed in Graph-Mamba for ObjectNav?}}

In this work, we investigate how to adapt the node prioritization mechanism in Graph-Mamba to object-goal navigation. 
We observe that object-goal navigation inherently provides a natural notion of importance: the relevance of each object to the target. 

Based on this observation, we propose Graph-MambaNav, a target-aware spatial-temporal graph encoding framework that orders object nodes according to their relevance to the target while modeling object dynamics over time. Unlike prior graph-based ObjectNav methods that encode target relevance only through features or attention weights, Graph-MambaNav explicitly turns target-object relevance into the computation order of graph sequence modeling, thereby making information propagation itself target-aware.

To obtain meaningful object relationships and ordering, we leverage Large Language Models (LLMs) ChatGPT5 to provide commonsense priors \cite{sun2025enhancing}. 
Specifically, we construct an affinity matrix as illustrated in Fig.~\ref{intro} that encodes pairwise object relationships conditioned on the target using carefully designed Chain-of-Thought \cite{wei2022chain} prompts (\textit{``Given typical indoor scenes, estimate the likelihood that object A appears within close spatial proximity (e.g., same functional area or within a few meters) of object B. The estimation should consider both statistical co-occurrence and functional relationships between objects based on your commonsense knowledge. Return a scalar score in the range [0, 1], where 0 indicates no association and 1 indicates a strong and consistent association''}). 
These LLM-derived priors are used to initialize both edge weights for local message passing and node ordering for global sequence modeling, enabling consistent and task-aligned information propagation across both local and global modules in proposed Graph-MambaNav.

Building on this prior, our spatial module integrates local message passing with global Mamba-based selective scanning. 
The local branch captures neighborhood-level interactions guided by learnable edge weights, while the global branch propagates information through a target-aware ordering sequence, enabling long-range reasoning across objects. 
Furthermore, since object navigation is inherently episodic and requires long-horizon sequential reasoning, we extend this framework to the temporal domain.
We model per-object temporal trajectories by applying a temporal SSMs module to object-aligned sequences across time, ensuring consistent tracking of each object while selectively aggregating historical observations into a memory representation.

We evaluate our approach on standard benchmarks, including AI2-THOR \cite{kolve2017ai2} and RoboTHOR \cite{deitke2020robothor}, and further validate its effectiveness on a real robotic platform. 
Experimental results demonstrate that our method improves navigation performance in simulation and generalizes effectively to unseen real-world environments, particularly for long-horizon episodes.

\textbf{Contributions}
The main contributions of this work are summarized as follows:
\begin{itemize}
    \item We propose Graph-MambaNav, a target-aware spatial-temporal graph encoding framework for object-goal navigation that enables fine-grained, computation-level control over object-to-target information propagation via efficient state space modeling.

   \item We investigate the node prioritization mechanism in Graph-Mamba and redesign it for object navigation by introducing LLM-derived node ordering and edge weights, where both object-to-target importance and pairwise relationships are jointly defined based on commonsense knowledge, yielding a task-driven formulation of graph structure and computation order.

    \item We demonstrate that the proposed framework achieves improved navigation performance in simulation compared to other graph-based end-to-end methods, and generalizes effectively to real-world robotic deployment in unseen indoor environments.
\end{itemize}


\begin{figure*}[t]
\centering
\includegraphics[width=\textwidth]{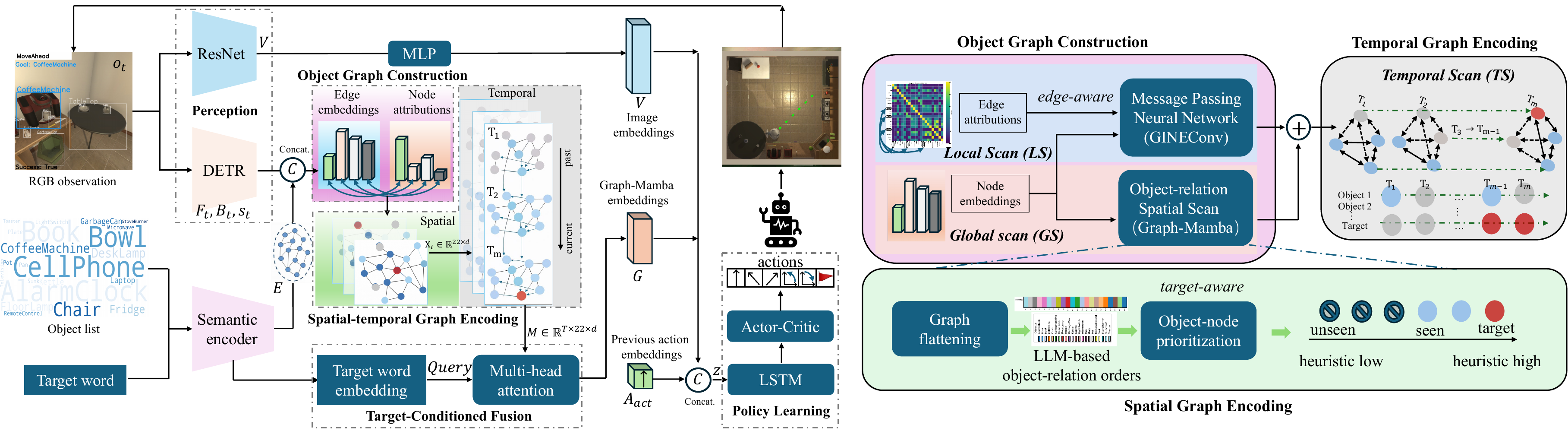}
\caption{
Overview of Graph-MambaNav (Left) and its object-graph spatial-temporal encoding pipeline (Right).
Given an egocentric RGB observation $\mathbf{o}_t$, ResNet extracts global scene-level features $\mathbf{V}$, while DETR provides complementary object-level features, boxes, and detection scores for object graph construction.
LLM-derived commonsense priors initialize edge embeddings and define target-aware node ordering. 
The enlarged module on the right shows the three key graph encoding operations: local edge-aware message passing, global target-aware Graph-Mamba scanning based on object-node prioritization, and temporal scanning over object-wise memory. 
The target-conditioned attention fusion module produces graph feature $\mathbf{G}$, which is fused with $\mathbf{V}$ and previous-action embedding $\mathbf{A}_{\mathrm{act}}$ for LSTM-based A3C policy learning.
}
\label{pipeline}
\vspace{-0.6cm}
\end{figure*}

\section{Related Works}
\subsection{State Space Models}
State Space Models (SSMs) \cite{gu2021efficiently} provide a principled approach to sequence modeling by describing the evolution of a latent state over time. 
At each step, the model updates its hidden state based on the current input and previously accumulated context, enabling efficient propagation of historical information. 
Unlike attention mechanisms that explicitly compute interactions between all pairs of tokens, SSMs summarize past observations into a compact state representation.
Advances in structured SSMs have improved both stability and scalability, making them competitive with modern deep sequence models. 
In contrast to Transformer-based architectures with quadratic complexity, SSMs achieve linear-time computation and reduced memory consumption \cite{sun2025memory}, which is particularly advantageous for long-horizon scenarios \cite{li2024stg}.

Building upon structured state space models, Mamba \cite{gu2024mamba} introduces a data-dependent selection mechanism for adaptive information propagation. 
It selectively updates hidden states to model long-range dependencies with linear-time efficiency. 
These properties make Mamba particularly attractive for embodied AI and robotics \cite{liu2024robomamba}, where agents must reason over long-horizon observations under limited memory \cite{zhou2025mtil}. This motivates its extension to graph-structured data, resulting in Graph-Mamba~\cite{wang2024graph}. Nevertheless, adapting Graph-Mamba to robotics remains non-trivial, especially in task-driven scenarios where node ordering and edge attributions must reflect task-specific priors, such as object relationships in ObjectNav.

\subsection{Graph-based Object-goal Navigation}
Graph-based representations have been widely explored to model object correlations for navigation. 
Du et al.~\cite{du2020learning} construct object relation graphs from detection results, while Zhang et al.~\cite{zhang2021hierarchical} introduce hierarchical object-to-zone graphs to encode high-level semantic priors. 
Dang et al.~\cite{dang2022unbiased} propose directed object attention for asymmetric interactions, and Zhang et al.~\cite{zhang2023layout} learn object layout distributions to capture spatial correlations. 
These works demonstrate that modeling object relationships improves navigation efficiency and generalization.

To leverage such structures, prior methods typically employ graph neural networks to propagate information over object graphs. 
Yang et al.~\cite{yang2018visual} use GNNs for co-occurrence modeling, Kwon et al.~\cite{kwon2021visual} introduce graph memory for temporal context, and Moghaddam et al.~\cite{moghaddam2021optimistic} adopt Graph Transformers for global reasoning. 
More recent approaches further enhance graph representations with richer features and temporal modeling. 
Chen et al.~\cite{chen2025temporal} propose TSOG to jointly model local object-object and global object-observation correlations with temporal aggregation, while Meng et al.~\cite{meng2025context} introduce context-aware graph inference with dynamic relationships conditioned on image, action, and memory.


However, existing graph-based ObjectNav methods, including CGI\cite{meng2025context}, AKGVP\cite{xu2024aligning} and TSOG\cite{chen2025temporal}, mainly encode target relevance through feature fusion, attention, or temporal aggregation over graph representations, while the graph computation itself remains largely unordered or permutation-invariant. In contrast, Graph-MambaNav does not simply replace the graph-processing module with Graph-Mamba. It redesigns Graph-Mamba’s node-prioritization mechanism for ObjectNav by using target-object relevance to determine the scan order and by using the same LLM-derived object-relation prior to initialize edge attribution, thereby making both graph structure and computation order target-aware.

\section{Task Definition}

We consider the object-goal navigation task, where an agent is required to locate a specified target object $o^\star \in \mathcal{O}$ within an environment $q \in \mathcal{Q}$.
At the beginning of each episode, a target object is randomly sampled, and the agent is initialized at a random pose $(x, y, \theta_{\text{yaw}}, \theta_{\text{pitch}})$ in a randomly selected environment. 
Here, $(x, y)$ denotes the agent's position on the ground plane, while $\theta_{\text{yaw}}$ and $\theta_{\text{pitch}}$ represent its orientation.

During navigation, the agent operates under partial observability and only receives a single egocentric RGB observation $\mathbf{o}_t$ at each time step. 
No global map or prior knowledge of the environment layout is provided. 
Based on the current observation and the target object, the agent learns a policy $\pi(a_t \mid \mathbf{o}_t, o^\star)$ to select an action $a_t$ from a discrete action space: \{\texttt{MoveAhead}, \texttt{RotateLeft}, \texttt{RotateRight},
\texttt{LookUp}, \texttt{LookDown}, \texttt{Done}\}.

Specifically, \texttt{MoveAhead} moves the agent forward by a fixed step, while \texttt{RotateLeft} and \texttt{RotateRight} adjust the agent’s orientation horizontally. 
\texttt{LookUp} and \texttt{LookDown} control the vertical viewing angle. 
The \texttt{Done} action is used to indicate that the agent believes it has reached the target.

An episode is considered successful only when the agent activates the \texttt{Done} action within a predefined step limit and, at termination, satisfies both of the following conditions: 
(1) the target object is visible in the agent's current field of view, and 
(2) the distance between the agent and the target is below a predefined threshold (e.g., 1.5\,m). 
Otherwise, the episode is regarded as a failure.

\section{Proposed Method}
\begin{algorithm}[t]

\caption{Target-aware Spatial-Temporal Graph Encoding with Object-Centric SSMs Scanning}
\label{alg:tsog}
\begin{algorithmic}[1]

\Require
Object features $\mathbf{F}_t \in \mathbb{R}^{22\times256}$ at time $t$,
text embeddings $\mathbf{E} \in \mathbb{R}^{22\times300}$,
detection info $(\mathbf{B}_t,\mathbf{s}_t)$,
target object $o^\star$,
affinity matrix $\mathbf{A}\in\mathbb{R}^{22\times22}$,
temporal memory $\mathbf{M}\in\mathbb{R}^{T\times22\times d}$

\Ensure
Updated memory $\mathbf{M}$ and spatial embeddings $\mathbf{X}_t$

\vspace{0.5em}
\Statex \textbf{Spatial Graph Construction}
\State $i^\star \leftarrow \texttt{index}(o^\star)$
\State $m_i \leftarrow (\|\mathbf{F}_t^{(i)}\|_1 > 0)$
\State $\mathbf{Z}_t \leftarrow \texttt{concat}(\mathbf{E},\mathbf{F}_t,\mathbf{B}_t,\mathbf{s}_t)$
\State $\mathbf{x} \leftarrow \mathbf{W}_{\text{node}}\cdot\texttt{concat}(\mathbf{Z}_t,\mathbf{1}_{i^\star})$

\State $\mathcal{E} \leftarrow \{(i,j)\mid m_i=1,\ m_j=1,\ i\neq j\}$
\State $e_{ij} \leftarrow \phi_{\text{edge}}(\mathbf{A}[i,j])$

\vspace{0.5em}
\Statex \textbf{Target-aware Spatial Graph-Mamba}
\State $h_i \leftarrow \mathbf{A}[i,i^\star]$
\State $h_{i^\star}\leftarrow +\infty$
\State $\rho \leftarrow \texttt{argsort}(h)$
\State $\rho^{-1} \leftarrow \texttt{inverse}(\rho)$

\For{$\ell = 1$ to $L_s$}
    \State $\mathbf{x}_{\text{loc}} \leftarrow \texttt{GINE}_\ell(\mathbf{x},\mathcal{E},e_{ij})$
    \State $\mathbf{x}_{\text{glo}} \leftarrow \texttt{MambaNode}_\ell(\mathbf{x}[\rho])[\rho^{-1}]$
    \State $\mathbf{x} \leftarrow \texttt{FFN}_\ell(\mathbf{x}_{\text{loc}} + \mathbf{x}_{\text{glo}})$
\EndFor
\State $\mathbf{X}_t \leftarrow \mathbf{x}$

\vspace{0.5em}
\Statex \textbf{Object-track Temporal Mamba}
\State $\mathbf{M} \leftarrow \texttt{PushBack}(\mathbf{M}, \mathbf{X}_t)$
\State $\mathbf{U} \leftarrow \texttt{Permute}(\mathbf{M})$

\For{$\ell = 1$ to $L_t$}
    \State $\mathbf{U} \leftarrow \texttt{MambaTime}_\ell(\mathbf{U})$
\EndFor

\State $\mathbf{M} \leftarrow \texttt{Permute}^{-1}(\mathbf{U})$

\State \Return $\mathbf{M}, \mathbf{X}_t$

\end{algorithmic}
\end{algorithm}

\subsection{Spatial Graph Modeling with Local-Global Fusion}


At each time step $t$, we model object-level spatial dependencies with a hybrid local-global graph module, as illustrated in Fig.~\ref{pipeline}. We use a COCO-pretrained DETR model~\cite{carion2020end} as a frozen object detector to extract object-level observations from the current RGB frame. Specifically, DETR provides object appearance features, bounding boxes, and detection confidence scores, which are mapped to the predefined object categories and used to construct the object-centric graph. When multiple instances of the same category are detected, we use the highest-confidence instance. For undetected categories, DETR-based appearance and box features are set to zero; when no object is detected in a frame, all object nodes use zero-valued detection features while global visual features and temporal memory are still used for policy learning.

The input consists of object appearance features $\mathbf{F}_t \in \mathbb{R}^{22\times256}$, text embeddings $\mathbf{E} \in \mathbb{R}^{22\times300}$, detection boxes $\mathbf{B}_t \in \mathbb{R}^{22\times4}$, and detection scores $\mathbf{s}_t \in \mathbb{R}^{22\times1}$. The semantic encoder in Fig. \ref{pipeline} denotes the GloVe-based \cite{pennington2014glove} object text embedding module, which maps each object category name into a 300-dimensional semantic embedding space. These features are concatenated to form $\mathbf{Z}_t \in \mathbb{R}^{22\times561}$, which is further augmented with a target indicator $\mathbf{1}_{i^\star}$ and projected into the initial latent node space as $\mathbf{x}^{(0)} \in \mathbb{R}^{22\times d}$.

\paragraph{Local graph modeling}
We construct a spatial graph over observed objects with edge set $\mathcal{E}=\{(i,j)\mid m_i=1,\;m_j=1,\;i\neq j\}$, where $m_i\in\{0,1\}$ indicates whether object category $i$ is observed in the current frame.
Each edge is assigned a prior score from an affinity matrix $\mathbf{A}$, initialized from LLM-derived commonsense relations and mapped into a learnable embedding $e_{ij}=\phi_{\text{edge}}(\mathbf{A}[i,j])$.

Based on these node and edge features, the local branch applies a GINE-based message passing layer. At layer $\ell$, the node features are updated as:
\begin{equation}
\mathbf{x}^{(\ell)}_{\text{loc}}
=
\mathrm{GINE}_{\ell}\!\left(
\mathbf{x}^{(\ell-1)},\mathcal{E},\{e_{ij}\}
\right)
\in\mathbb{R}^{22\times d},
\end{equation}
which aggregates neighborhood information with explicit edge-aware relational encoding.

\paragraph{Global graph modeling}
To capture long-range dependencies, we introduce a target-aware global scan inspired by Graph-Mamba. 
We define a heuristic ordering $h_i=\mathbf{A}[i,i^\star]$ (with $h_{i^\star}=+\infty$), such that nodes are sorted by relevance to the target.
This orders nodes from low to high target relevance, with the target placed last to aggregate richer causal context under causal scanning.

The reordered node embeddings are processed by a Mamba-based sequence model:
\begin{equation}
\mathbf{x}^{(\ell)}_{\text{glo}}
=
\mathrm{MambaNode}_{\ell}\!\left(
\mathbf{x}^{(\ell-1)}[\rho]
\right)[\rho^{-1}]
\in\mathbb{R}^{22\times d}.
\end{equation}
where $\rho$ is the permutation that sorts nodes by target relevance, and $\rho^{-1}$ restores the original order.


\paragraph{Local-global fusion.}
The local and global branches are combined at each layer as
\begin{equation}
\mathbf{x}^{(\ell)}
=
\mathrm{FFN}_{\ell}\!\left(
\mathbf{x}^{(\ell)}_{\text{loc}}+\mathbf{x}^{(\ell)}_{\text{glo}}
\right),
\end{equation}
and stacked over $L_s$ layers to produce the final spatial representation $\mathbf{X}_t=\mathbf{x}^{(L_s)}$.


Overall, the local branch captures edge-aware object interactions, while the global branch models target-aware long-range dependencies. These two branches are complementary: the local branch grounds relational reasoning in observed structure, whereas the global branch enables target-guided information propagation across distant objects, jointly yielding a spatial representation tailored for goal-directed navigation.

\subsection{Temporal Object-Track Sequence Modeling}

Given the spatial embeddings $\mathbf{X}_t \in \mathbb{R}^{22\times d}$ at time $t$, we model temporal dependencies at the object level by maintaining a memory tensor $\mathbf{M} \in \mathbb{R}^{T \times 22 \times d}$, where each slice corresponds to a time step and each column corresponds to a fixed object identity. 
At each step, the current embeddings are appended via a First-In First-Out (FIFO) update $\mathbf{M} \leftarrow \texttt{PushBack}(\mathbf{M}, \mathbf{X}_t)$, maintaining the most recent $T$ observations in chronological order.

To enable object-centric temporal modeling, we reinterpret the memory as a set of object tracks by permuting the dimensions, i.e., $\mathbf{U} = \texttt{Permute}(\mathbf{M}) \in \mathbb{R}^{22 \times T \times d}$, where each sequence $\mathbf{U}_i \in \mathbb{R}^{T \times d}$ represents the temporal evolution of object $i$. The object order remains fixed across time, ensuring consistent identity alignment.


We then apply a Mamba-based sequence model along the temporal dimension:
\begin{equation}
\mathbf{U} \leftarrow \mathrm{MambaTime}_\ell(\mathbf{U}), \quad \ell = 1, \dots, L_t,
\end{equation}
where $L_t$ denotes the number of temporal Mamba-based object-track sequence layers. 
This module performs causal selective scanning to aggregate long-range temporal context. Finally, the updated representations are mapped back to the original layout via $\mathbf{M} = \texttt{Permute}^{-1}(\mathbf{U}) \in \mathbb{R}^{T \times 22 \times d}$.

Unlike the spatial module, which models inter-object relations at a single time step, the temporal module captures the evolution of each object independently across time, enabling consistent object-wise temporal reasoning and long-horizon dependency modeling.

\subsection{Target-Conditioned Spatial-Temporal Fusion}
Given the temporally updated memory $\mathbf{M} \in \mathbb{R}^{T \times 22 \times d}$, we extract target-relevant information via a target-conditioned attention mechanism. 
The target embedding is obtained from text features as $\mathbf{q} = \mathbf{W}_{\text{text}} \cdot \mathbf{e}_{i^\star}$. Here, $i^\ast$ is the target object index, $\mathbf{e}_{i^\ast}\in\mathbb{R}^{300}$ denotes its text embedding, and $\mathbf{W}_{\text{text}}$ projects it into the 256-dimensional attention hidden space to form the target query $\mathbf{q}$.

We then apply multi-head cross-attention, where the target embedding serves as the query and the temporal memory as key-value pairs \cite{chen2025temporal} \cite{sun2025memory}:
\begin{equation}
\mathbf{H} = \mathrm{Attention}(\mathbf{q}, \mathbf{M}, \mathbf{M}),
\end{equation}
which selectively aggregates target-relevant temporal context.


\begin{figure*}[t]
\centering
\includegraphics[width=0.98\textwidth]{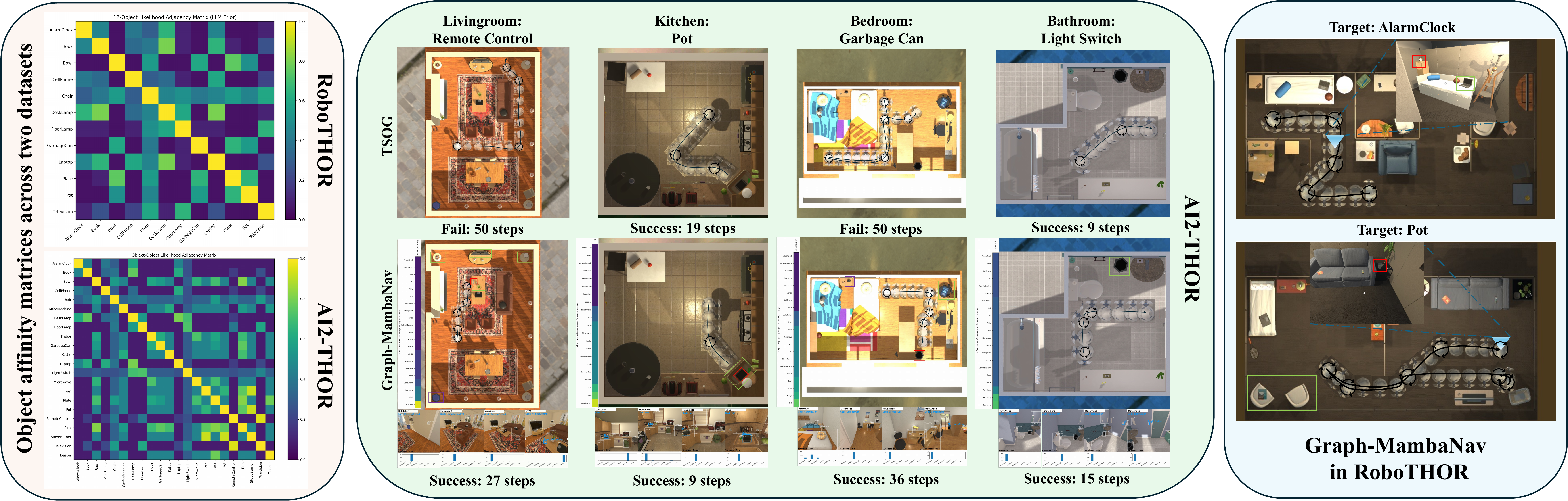}
\caption{
Left: LLM-derived object affinity matrices across two datasets.
Middle: Qualitative comparison of TSOG \cite{chen2025temporal} and Graph-MambaNav in AI2-THOR, covering four different room types.
Right: Graph-MambaNav results in RoboTHOR, where apartments are constructed modularly from a shared asset library.
Object-relation heatmaps for the target are shown to the left of the top-down maps.
In the top-down views, targets are highlighted in red and related objects in green, with the corresponding first-person observations shown below. 
}
\label{demo}
\vspace{-0.6cm}
\end{figure*}
The attended representation $H\in\mathbb{R}^{1\times22\times256}$, containing 22 object tokens, is projected by a shared MLP $256\rightarrow128\rightarrow49$ and reshaped to $G\in\mathbb{R}^{1\times22\times7\times7}$.
In parallel, the global visual feature $\mathbf{V}$ is extracted from the current RGB observation using a pretrained ResNet-18 backbone~\cite{he2016deep}, and the previous action is encoded as $\mathbf{A}_{\text{act}}$ with the same spatial resolution as $\mathbf{G}$, with the previous-action input initialized as a zero vector at the first timestep. The three modalities are fused via channel-wise concatenation $\mathbf{F} = \texttt{concat}(\mathbf{V}, \mathbf{G}, \mathbf{A}_{\text{act}})\in\mathbb{R}^{1\times96\times7\times7}$, followed by a point-wise convolution to produce the final representation $\mathbf{z} = \mathrm{Conv}_{1\times1}(\mathbf{F})$. This design enables the policy to jointly leverage visual observations, spatial-temporal memory, and action history, while the target-conditioned attention acts as a filtering mechanism to prioritize task-relevant object dynamics.

\subsection{Policy Learning}


The fused representation $\mathbf{z}$, obtained from the spatial-temporal fusion module, is used as the input to the policy network. 
To model sequential dependencies in decision making, $\mathbf{z}$ is fed into a recurrent module \cite{chen2025temporal,meng2025context,sun2025memory}, producing the recurrent hidden state $\mathbf{h}_t$.
We formulate the task as a sequential decision-making problem and optimize the policy using the Asynchronous Advantage Actor-Critic (A3C) framework \cite{mnih2016asynchronous}. 
The policy head produces an action distribution $\pi(a_t \mid \mathbf{h}_t)$, while a value head estimates the state value $V(\mathbf{h}_t)$ for advantage-based policy learning.



\section{Experiments}
\subsection{Datasets}
We evaluate the generalization capability of the proposed method in unseen environments using the AI2-THOR~\cite{kolve2017ai2} and RoboTHOR~\cite{deitke2020robothor} simulators. 
AI2-THOR consists of 120 indoor scenes spanning four categories: kitchens, living rooms, bedrooms, and bathrooms, each differing in layout and object arrangements. 
Following standard protocols~\cite{zhang2023layout,xu2024aligning,chen2025temporal}, we partition the scenes of each category into 20 for training, 5 for validation, and 5 for testing.

We consider 22 object categories as navigation targets, ensuring that each scene type contains at least four target instances. 
The selected categories include common household objects such as AlarmClock, Book, Bowl, CellPhone, Chair, CoffeeMachine, DeskLamp, FloorLamp, Fridge, GarbageCan, Kettle, Laptop, LightSwitch, Microwave, Pan, Plate, Pot, RemoteControl, Sink, StoveBurner, Television, and Toaster.

RoboTHOR contains 89 apartment environments, with larger spatial extent and longer navigation trajectories compared to AI2-THOR. 
Following prior works~\cite{zhang2021hierarchical,zhang2023layout,chen2025temporal}, we split the environments into 60 for training, 5 for validation, and 10 for testing. 
We select 12 object categories as targets in RoboTHOR, including AlarmClock, Book, Bowl, CellPhone, Chair, DeskLamp, FloorLamp, GarbageCan, Laptop, Plate, Pot, and Television.
\subsection{Evaluation Metrics}
For fair comparison, we follow the standard evaluation protocols used in prior ObjectNav studies~\cite{chi2025navigating,chen2025temporal,meng2025context,sun2025memory} to assess the generalization performance of the proposed method in unseen environments. Specifically, we use the commonly adopted AI2-THOR/RoboTHOR splits, target object categories, and evaluation metrics, including Success Rate (SR) and Success weighted by Path Length (SPL). Except for Memory-MambaNav, the baseline results are taken from the corresponding original papers under the same or comparable evaluation settings. For Memory-MambaNav\cite{sun2025memory}, we use our own reimplementation without its additional reward-design component, so that the comparison focuses on the modeling difference between vanilla Mamba-based memory and the proposed graph Mamba-based spatial-temporal reasoning.

For evaluation, each result is obtained from three repeated test runs under the same protocol, and we report the mean and standard deviation. The success rate (SR) measures the fraction of successful episodes, defined as $\text{SR} = \frac{1}{N} \sum_{i=1}^{N} S_i$, where $N$ is the total number of episodes and $S_i \in \{0,1\}$ indicates whether the $i$-th episode is successful. SPL further evaluates navigation efficiency by comparing the executed trajectory length $P_i$ with the shortest-path distance $L_i$, defined as $\text{SPL} = \frac{1}{N} \sum_{i=1}^{N} S_i \frac{L_i}{\max(L_i, P_i)}$. We report results over all trajectories (ALL) as well as a subset of more challenging cases where the shortest-path distance satisfies $L_i \geq 5$.

\subsection{Implementation Details}

\begin{figure}[t]
\centering
\includegraphics[width=0.32\textwidth]{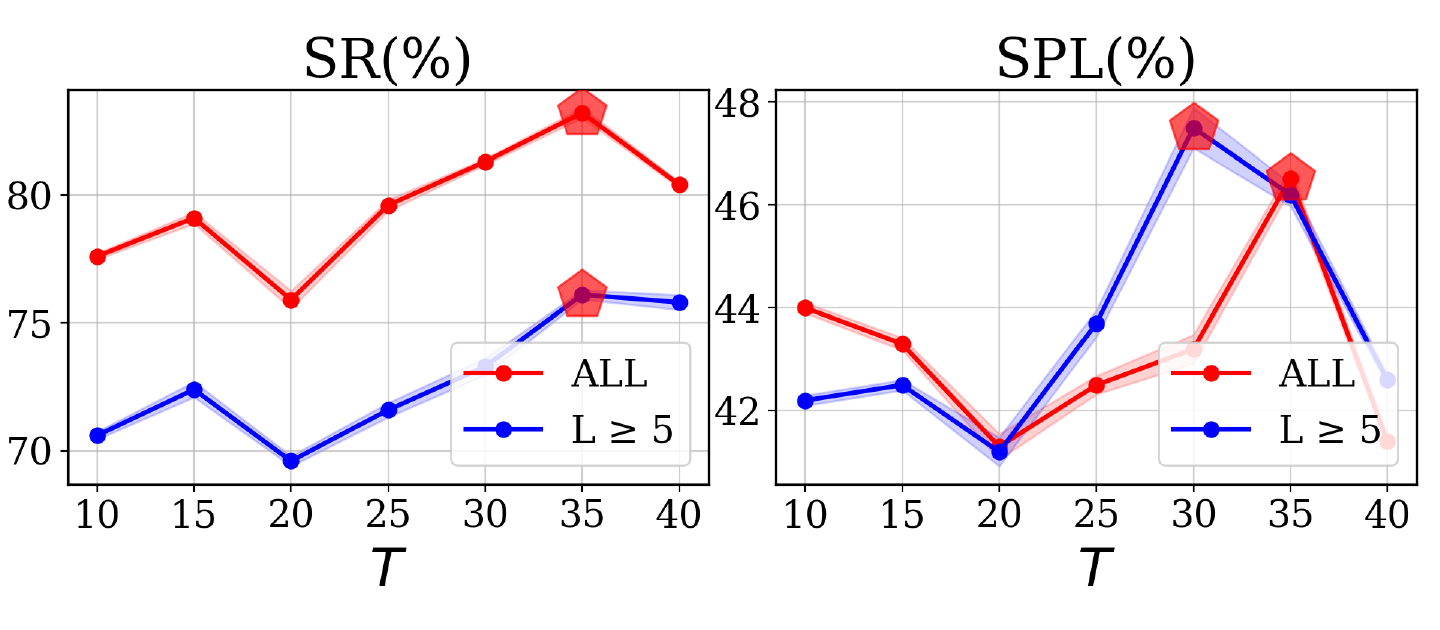}
\caption{Navigation performance across varying $T$. Red pentagon markers denote the optimal configurations.}
\label{m-len}
\vspace{-0.6cm}
\end{figure}

We train the proposed model using the A3C framework for several million episodes on three RTX 3080Ti GPUs. At each step, a small negative reward ($-0.01$) is applied to encourage efficiency, while a positive reward (5) is given upon successful completion. 
To promote exploration, an additional reward of $0.01$ is assigned when the agent executes the \texttt{MoveAhead} action. The model is optimized using Adam with a learning rate of $10^{-4}$, a discount factor of $0.99$, and a value loss coefficient of $0.5$. 
Unless otherwise specified, all projection layers are implemented as fully connected linear layers with bias. For MLP projections, ReLU activation is used between consecutive linear layers. The policy network uses a two-layer LSTM with 512 hidden units. The object and graph feature dimensions are 64 and 256, respectively, and the temporal memory length is capped at $T = 35$. 
As shown in Fig.~\ref{m-len}, SR and SPL increase with $T$ and then decline. 
The best trade-off is achieved at $T = 35$, balancing informative context and memory noise in the proposed framework.

To substantiate the computational efficiency of the proposed Mamba-based design, we further compare T35-Graph-MambaNav with an alternative Transformer-based variant under the same evaluation setting. 
As shown in Table~\ref{tab:computational_complexity}, T35-Graph-MambaNav requires fewer parameters, lower GFLOPs, and less GPU memory with one evaluation worker, while achieving a higher FPS. 
This indicates that the proposed graph Mamba-based sequence modeling provides a more efficient alternative to Transformer-based sequence modeling for long-horizon object-goal navigation.

\begin{table}[t]
\centering
\caption{
Comparison of computational complexity. 
We report the floating point operations (FLOPs), number of model parameters, FPS, and GPU memory consumption of the proposed T35-Graph-MambaNav and the alternative Transformer-based variant, both using a memory sequence length of 35.
}
\label{tab:computational_complexity}
\footnotesize
\resizebox{\linewidth}{!}{
\begin{tabular}{lcccc}
\hline
Method & Params & GFLOPs & GPU Memory & FPS \\
       & (M)    &        & (GB) per agent          &     \\
\hline
T35-Transformer variant & 62.13 & 82.39 & 6.72 & 11.32 \\
\rowcolor{blue!20} T35-Graph-MambaNav & 12.58 & 23.47 & 2.53      & 22.12 \\
\hline
\end{tabular}
}
\end{table}

\begin{table}[t]
\vspace{2mm}
\centering
\caption{Comparisons with other methods in the AI2-THOR and RoboTHOR environments}
\label{tab:comparison}
\resizebox{0.48\textwidth}{!}{
\begin{tabular}{l|cc|cc||cc|cc}
\hline
\multirow{2}{*}{Method} 
& \multicolumn{4}{c||}{AI2-THOR} 
& \multicolumn{4}{c}{RoboTHOR} \\
\cline{2-9}
& \multicolumn{2}{c|}{ALL} 
& \multicolumn{2}{c||}{$L \geq 5$}
& \multicolumn{2}{c|}{ALL}
& \multicolumn{2}{c}{$L \geq 5$} \\
\cline{2-9}
& SR$\uparrow$(\%) & SPL$\uparrow$(\%) 
& SR$\uparrow$(\%) & SPL$\uparrow$(\%)
& SR$\uparrow$(\%) & SPL$\uparrow$(\%)
& SR$\uparrow$(\%) & SPL$\uparrow$(\%) \\
\hline
Random          & 3.56  & 1.73  & 0.27  & 0.07  & -  & -  & -  & - \\
Baseline \cite{zhu2017target} & 57.35 & 33.78 & 45.77 & 30.65 & 26.41 & 16.61 & 17.42 & 12.23 \\
SP  \cite{yang2018visual}        & 62.16 & 37.01 & 50.86 & 34.17 & 28.04 & 17.63 & 21.66 & 15.14 \\
SAVN  \cite{wortsman2019learning}       & 63.32 & 37.62 & 52.38 & 35.31 & 28.42 & 17.82 & 22.13 & 15.34 \\
SpAtt \cite{mayo2021visual}      & 65.61 & 38.93 & 54.11 & 35.89 & 28.53 & 18.27 & 22.35 & 15.45 \\
ORG    \cite{du2020learning}     & 66.38 & 38.42 & 55.55 & 36.26 & 29.61 & 19.23 & 22.53 & 15.73 \\
ORG+TPN  \cite{du2020learning}   & 67.31 & 39.53 & 57.41 & 38.27 & 30.01 & 20.51 & 22.25 & 16.64 \\
HOZ   \cite{zhang2021hierarchical}      & 70.62 & 40.02 & 62.75 & 39.24 & 32.27 & 20.48 & 24.83 & 16.89 \\
VTNet   \cite{du2021vtnet}   & 72.20 & 44.90 & 63.40 & 44.00 & 31.62 & 19.63 & 23.48 & 16.02 \\
L-sTDE  \cite{zhang2023layout}   & 74.19 & 40.30 & 64.01 & 39.97 & 42.13 & 24.54 & 32.04 & 17.44 \\
AKGVP  \cite{xu2024aligning}     & 73.63 & 40.66 & 63.51 & 39.63 & 39.69 & 25.84 & 28.55 & 18.79 \\
AKGVP-CI \cite{xu2024aligning}   & 76.78 & 39.63 & 65.45 & 39.01 & 44.53 & 27.61 & 32.68 & 20.55 \\

CGI+GAIL \cite{meng2025context}   & {77.59} & {46.25} & {69.18} & {46.10}& {48.30} & \textbf{28.77}& {36.69} & {21.89} \\
TSOG \cite{chen2025temporal}   & {80.04} & {41.44} & {73.46} & {43.66}& {-} & {-}& {-} & {-} \\                
M-MambaNav \cite{sun2025memory}  & {81.24} & \textbf{46.60} & {74.32} & {44.81}& {-} & {-} & {-} & {-} \\
\hline
\rowcolor{blue!20}
\textbf{Graph-MambaNav}   
& \textbf{83.22} & 46.52 & \textbf{76.09} & \textbf{46.20}
& \textbf{49.82} & 28.67 & \textbf{37.38} & \textbf{22.49} \\
\rowcolor{gray!8}
\scriptsize{Standard deviation}
& \scriptsize{$\pm$0.75} & \scriptsize{$\pm$0.61}
& \scriptsize{$\pm$0.92} & \scriptsize{$\pm$0.72}
& \scriptsize{$\pm$0.66} & \scriptsize{$\pm$0.63}
& \scriptsize{$\pm$1.12} & \scriptsize{$\pm$0.74} \\
\hline
\end{tabular}
}
\vspace{-0.4cm}
\end{table}

\begin{figure}[t]
\centering
\includegraphics[width=0.36\textwidth]{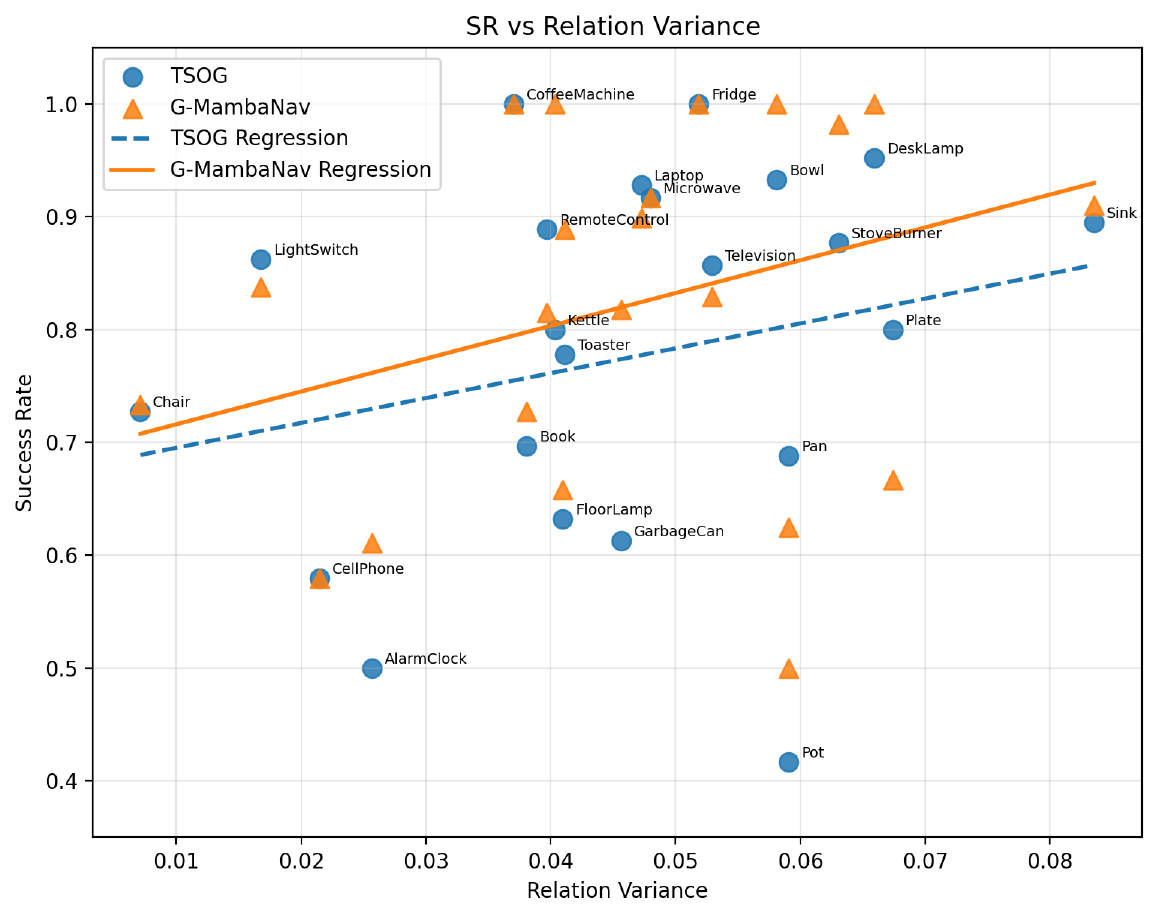}
\caption{SR versus relation variance of each target object
        between Graph-MambaNav and TSOG. Graph-MambaNav shows a steeper trend and higher SR, demonstrating its advantage in leveraging prior LLM-derived relational knowledge.}
\label{discriminability_analysis}
\vspace{-0.4cm}
\end{figure}


\subsection{Results}
\subsubsection{Qualitative Results}

We visualize navigation trajectories in both AI2-THOR \cite{kolve2017ai2} and RoboTHOR \cite{deitke2020robothor} and compare our method with recent method TSOG \cite{chen2025temporal}, as shown in Fig.~\ref{demo}. 
Overall, the results validate the effectiveness of our object-relation modeling in guiding target-oriented exploration. In a \textit{living room} scenario where the agent searches for a \textit{remote control}, our method leverages object relationships to guide exploration. 
When observing semantically related objects such as \textit{laptop}, \textit{floor lamp}, and \textit{television}, the agent actively moves toward these regions, where the smaller target object is more likely to appear. 
Similar behaviors are observed in other scenes, such as locating a \textit{pot} near a \textit{sink} in a \textit{kitchen}, and finding a \textit{garbage can} in proximity to a \textit{chair} in a \textit{bedroom}. In these cases, our method demonstrates higher success rates and more efficient trajectories compared to TSOG. 
 However, for targets with less distinctive relational cues, such as \textit{light switch} with relative low relation variance as illustrated in Fig. \ref{discriminability_analysis}, the reliance on prior object relationships may introduce bias. 
For example, the agent tends to navigate toward objects with higher prior correlation (e.g., \textit{garbage can} in Bathroom case, which may reduce exploration diversity. 
Another phenomenon is observed in RoboTHOR when searching for \textit{pot}, where the actual object layout deviates from the commonsense affinity prior, resulting in suboptimal exploration behavior.



These results highlight both the advantages of leveraging structured object relationships for efficient navigation and the limitations when such priors are less discriminative. As shown in Fig.~\ref{discriminability_analysis}, SR generally increases with relation variance, indicating a positive correlation between relational discriminativeness and navigation performance. 
The regression lines are obtained by fitting a linear model between relation variance and SR, reflecting this overall trend. 
Notably, our method exhibits a steeper slope and consistently higher SR than TSOG, suggesting it better exploits informative object relations. 
However, this trend is not absolute, as certain targets (e.g., \textit{pot}) still exhibit low SR despite high variance, indicating that relation variance is not the sole factor affecting performance.

\subsubsection{Quantitative Results}
We further compare our method with recent state-of-the-art approaches, including CGI+GAIL~\cite{meng2025context}, TSOG~\cite{chen2025temporal}, and Memory-MambaNav~\cite{sun2025memory}, as summarized in Table~\ref{tab:comparison}. 
On AI2-THOR, our Graph-MambaNav achieves strong overall performance, reaching {83.22\%} SR and {46.52\%} SPL on all trajectories, outperforming CGI+GAIL (77.59\% / 46.25\%), TSOG (80.04\% / 41.44\%), and Memory-MambaNav (81.24\% SR). 
The improvement is more evident on long trajectories ($L \geq 5$), where our method achieves {76.09\%} SR, demonstrating stronger long-horizon reasoning ability. As shown in Table~\ref{tab:path_length}, all methods degrade as path length increases, while ours maintains advantages over all other methods.
Improvements are more evident on longer trajectories, and removing temporal scan leads to clear drops, highlighting the importance of long-horizon temporal modeling for ObjectNav.

\begin{table}[t]
\centering
\caption{Comparison under different path lengths (10, 15, and 20) in AI2-THOR, where \textit{TS} denotes Temporal Scan}
\label{tab:path_length}
\resizebox{0.68\columnwidth}{!}{
\begin{tabular}{l|cc}
\hline
\multirow{2}{*}{Method} 
& \multicolumn{2}{c}{$L \geq 10$ / $L \geq 15$ / $L \geq 20$} \\
& SR$\uparrow$(\%) 
& SPL$\uparrow$(\%) \\
\hline

VTNet 
& 42.5 / 29.6 / 10.9
& 29.1 / 18.5 / 6.3
\\

DOA 
& 54.88 / 25.18 / 10.91
& 37.15 / 17.83 / 8.35
\\


TSOG 
& 58.49 / 35.31 / 23.64
& 37.18 / 22.11 / 16.31
\\

M-MambaNav
& 60.23 / {42.35}/ 29.35
& 38.81 / 25.93 / 16.21
\\

\hline
Ours w/o TS
& 61.23 / 41.02 / 27.32
& 40.33 / 28.19 / 15.65 
\\

\rowcolor{blue!20}\textbf{Ours} 
& \textbf{65.92} / \textbf{42.63} / \textbf{33.40}
& \textbf{44.33} / \textbf{30.32} / \textbf{18.25}
\\
\rowcolor{gray!8}\scriptsize{Std. dev.}
& \scriptsize{$\pm$0.90 / $\pm$1.03 / $\pm$1.28}
& \scriptsize{$\pm$0.78 / $\pm$0.93 / $\pm$1.15}  

\\
\hline
\end{tabular}
}
\end{table}

From a mechanism perspective, the comparison reveals complementary strengths of different approaches. 
Compared with CGI+GAIL \cite{meng2025context}, which enhances navigation through context-aware graph inference and reward shaping, our method benefits from explicit temporal memory modeling, enabling more effective accumulation of historical context. 
Compared with TSOG \cite{chen2025temporal}, which captures temporal correlations via graph aggregation, our method further leverages Mamba-based sequence modeling to achieve more efficient long-range dependency modeling. 
Compared with Memory-MambaNav \cite{sun2025memory}, which employs Mamba for temporal modeling but relies on fixed-length memory and lacks explicit relational structure, our method demonstrates the advantage of integrating graph-structured reasoning for object navigation. 


On RoboTHOR, our method achieves the highest SR ({49.82\%}) and competitive SPL (28.67\%), and maintains consistent improvements on long trajectories ($L \geq 5$), reaching {37.38\%} SR and {22.49\%} SPL. 
Compared with AI2-THOR, the gains are relatively smaller, which we attribute to differences in environment characteristics, as shown the comparison in Fig. \ref{demo}. 
AI2-THOR contains more structured and semantically consistent object layouts, where relational reasoning and temporal memory are more beneficial. 
In contrast, RoboTHOR presents greater variability as illustrated in Fig. \ref{demo}, where navigation efficiency (SPL) suggests that further improvements may be achieved through better reward design and policy optimization.

We further analyze ordering scalability and noisy-prior robustness in Table~\ref{tab:ordering_noise}. 
With a fixed object vocabulary, increasing the target-aware ordered nodes improves performance from top-5 to top-10 and then saturates from top-15 to all-22, indicating diminishing gains once major relational cues are covered. 
When 10\%, 20\%, and 30\% of symmetric off-diagonal affinity entries are corrupted with random values from a uniform distribution in $[0,1]$ and used for both edge attribution and ordering, performance gradually decreases, suggesting moderate robustness to partial relation noise.
 

\begin{table}[t]
\centering
\caption{
Analysis of ordering scalability and robustness to noisy object-relation priors.
Due to the fixed object categories in the offline benchmark, we vary the number of nodes using target-aware ordering and perturb the LLM-derived affinity matrix to simulate noisy relations.
}
\label{tab:ordering_noise}
\resizebox{0.67\columnwidth}{!}{
\begin{tabular}{lcc}
\hline
Setting & SR$\uparrow$(\%) & SPL$\uparrow$(\%) \\
\hline
Top-5 ordered nodes  & $77.05{\pm}0.91$ & $43.10{\pm}0.76$ \\
Top-10 ordered nodes & $81.56{\pm}0.78$ & $45.65{\pm}0.64$ \\
Top-15 ordered nodes & $82.63{\pm}0.72$ & $46.32{\pm}0.59$ \\
Top-20 ordered nodes & $82.84{\pm}0.69$ & $46.30{\pm}0.57$ \\
\rowcolor{blue!20} All-22 ordered nodes & $\textbf{83.22}{\pm}0.75$ & $\textbf{46.52}{\pm}0.61$ \\
\hline
All-22 + 10\% noisy relations & $82.31{\pm}0.83$ & $45.51{\pm}0.71$ \\
All-22 + 20\% noisy relations & $80.46{\pm}0.96$ & $43.09{\pm}0.84$ \\
All-22 + 30\% noisy relations & $77.92{\pm}1.12$ & $41.71{\pm}0.93$ \\
\hline
\end{tabular}
}
\vspace{-0.4cm}
\end{table}


\subsection{Ablation Studies}
We analyze the roles of Local, Global, and Temporal Scans in Table~\ref{tab:ablation} (Group I). 
LS provides moderate gains (63.83$\rightarrow$66.15 SR), while GS yields a substantially larger improvement (66.15$\rightarrow$78.34 SR), highlighting the importance of ordered global propagation. 
TS further boosts performance (78.34$\rightarrow$83.22 SR), especially on long trajectories, by enabling persistent temporal memory. 
These results suggest that effective navigation relies on jointly modeling local structure, global order, and temporal context, with global ordering being the dominant factor among them.


We further examine the effects of edge initialization and computation order in Mamba-based sequence modeling (Group II). 
All variants in Group II retain the full LS+GS+TS pipeline and modify only the specified relation prior or scan order. 
Using fixed random LS edges (Ours w/ F-R LS edges) yields limited performance (66.26 SR), indicating that unstructured relations are insufficient. 
Replacing them with learnable random-initialized LS edges (Ours w/ L-R LS edges) significantly improves performance (77.59 SR), suggesting that relational patterns can be learned from data. 
Using fixed LLM-derived LS edges (Ours w/ F-LLM LS edges) achieves comparable performance (77.49 SR), highlighting the effectiveness of commonsense priors. 
However, applying global scan with random ordering (Ours w/ GS RO) results in a clear performance drop (73.67 SR), indicating that improper computation order harms global scan reasoning.

In addition, applying temporal scanning in a current-to-past order (Ours w/ C-P TS) achieves competitive performance (81.24 SR). 
However, compared with our full model, this ordering places the current observation at the beginning of the sequence, limiting its ability to aggregate full historical context under causal scanning. 
In contrast, our chronological design (past-to-current) enables the current state to aggregate richer temporal context, consistent with the node-prioritization principle in Graph-Mamba and our global graph modeling, leading to further performance gains.

\begin{table}[t]
\centering
\caption{Ablation study, impacts of different components. All variants in Group II retain the full LS+GS+TS pipeline and modify only the specified relation prior or scan order.}
\label{tab:ablation}
\resizebox{0.66\columnwidth}{!}{
\begin{tabular}{cc|ccc|cc}
\hline
\multirow{2}{*}{ID} & 
& \multicolumn{3}{c|}{Method} 
& \multicolumn{2}{c}{ALL / $L \geq 5$} \\
\multicolumn{2}{c|}{} & LS & GS & TS
& SR$\uparrow$(\%) & SPL$\uparrow$(\%)  \\
\hline\hline
\multirow{4}{*}{I} & 1 &  &  &  & 63.83 / 52.45 & {33.75} / 30.33  \\
                  & 2 & $\checkmark$ &  &  & 66.15 / 61.42 & 41.35 / 35.43  \\
                  & 3 & $\checkmark$ & $\checkmark$ &  & 78.34 / 71.66 & 43.39 / 42.85  \\
                  \rowcolor{blue!20}& 4 & $\checkmark$ & $\checkmark$ & $\checkmark$
                    & \textbf{83.22} / \textbf{76.09} & \textbf{46.52} / \textbf{46.20}  \\
\hline\hline
\multirow{6}{*}{II} & 5 & \multicolumn{3}{c|}{Ours w/ F-R LS edges}
                    & 66.26 / 62.98 & 40.34 / 38.23  \\
                    & 6 & \multicolumn{3}{c|}{Ours w/ L-R LS edges}
                    & 77.59 / 68.81 & 45.27 / 42.49  \\
                     & 7 & \multicolumn{3}{c|}{Ours w/ F-LLM LS edges}
                    & 77.49 / 72.31 & 44.61 / 44.66  \\
                   & 8 & \multicolumn{3}{c|}{Ours w/ RO GS }
                    & 73.67 / 70.74 & {41.39} / 41.05  \\
                            & 9 & \multicolumn{3}{c|}{Ours w/ C-P TS }
                    & 81.24 / 73.46 & {43.82} / 42.03  \\
                   \rowcolor{blue!20}& 10 & \multicolumn{3}{c|}{\textbf{Ours (full)}}
                    & \textbf{83.22} / \textbf{76.09} & \textbf{46.52} / \textbf{46.20}  \\
\hline
\end{tabular}
}
\vspace{-0.4cm}
\end{table}

\subsection{Real-world Deployment}

\begin{figure}[t]
\centering
\includegraphics[width=0.48\textwidth]{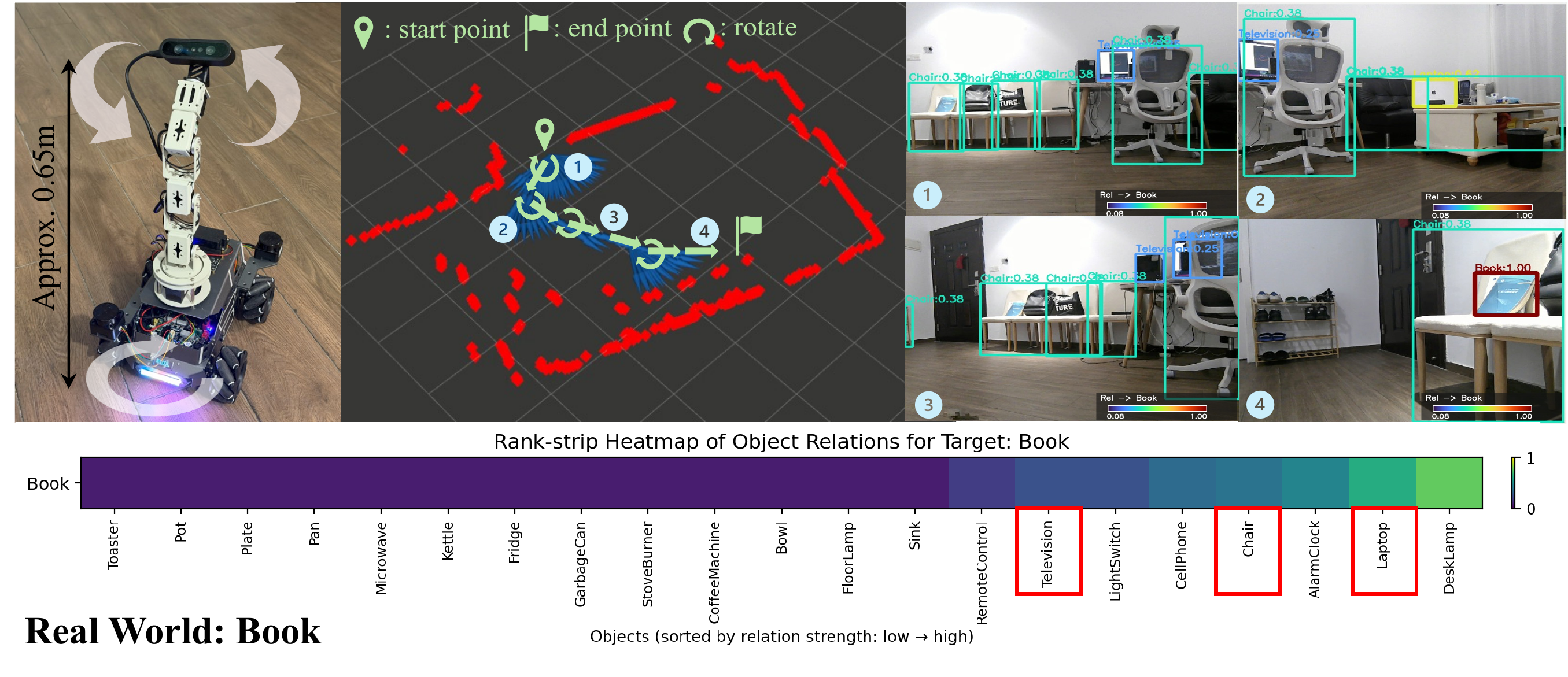}
\caption{
Real-world navigation illustration. Left: robot platform; middle: trajectory visualization; right: perception results. Bottom: object relation heatmap for the target \textit{book}.
The point cloud and odometry are for visualization only and are not accessible to the robot.
}
\label{realexp}
\vspace{-0.6cm}
\end{figure}

We develop a real-world system to evaluate the transferability of our method. 
The system is deployed on a wheeled robot with a monocular RGB camera mounted at 0.65\,m as illustrated in Fig \ref{realexp}, to closely match the simulation setup. 
Locomotion is controlled by the mobile base, while camera pitch is actuated by a servo. A distributed architecture is adopted: a Raspberry Pi 5 onboard runs ROS2 for low-level control, while the navigation model is executed on an external host. 
Both systems communicate via ROS2 topics over a local network for real-time perception and control.

We evaluate the system in an indoor living-room environment, where the target object is a \textit{book}. 
Objects with strong semantic relations to the target, such as \textit{laptop}, \textit{chair}, and \textit{television}, are present in the scene. 
During navigation, the robot first observes a \textit{laptop} and moves toward that region. 
After failing to detect the target, it continues exploring nearby related areas, including chairs and the television region, and finally navigates to a cluster of chairs where the \textit{book} is successfully located. The results demonstrate that our method can effectively transfer from simulation to real-world environments and leverage object relationships to guide exploration toward target-relevant regions.

\section{Conclusion}

In this work, we propose Graph-MambaNav, a target-aware spatial-temporal graph encoding framework for object-goal navigation. 
By integrating graph-structured relational reasoning with Mamba-based sequence modeling, our method enables ordered information propagation and effective long-horizon temporal modeling. 
Extensive experiments in both simulation and real-world environments demonstrate consistent improvements over existing methods. Our analysis further highlights the importance of aligning relational structure with computation order in graph-based SSMs selective scanning, as well as the impact of discriminative object relationships on object-goal navigation performance. Future work will extend Graph-MambaNav to larger or open-vocabulary object sets and model multiple instances per category beyond the current fixed object graph.


%




\section*{Acknowledgment}

Leyuan Sun would like to express his heartfelt gratitude to his wife for her unwavering support, understanding, and dedication to their family. This manuscript was submitted while they awaited the birth of their son, making its completion especially meaningful. Together, they wish their son a life filled with health, happiness, and boundless curiosity.

\ifCLASSOPTIONcaptionsoff
  \newpage
\fi

\bibliographystyle{IEEEtran}
\bibliography{RAL2026}

@article{gu2021efficiently,
  title={Efficiently modeling long sequences with structured state spaces},
  author={Gu, Albert and Goel, Karan and R{\'e}, Christopher},
  journal={arXiv preprint arXiv:2111.00396},
  year={2021}
}

@inproceedings{gu2024mamba,
  title={Mamba: Linear-time sequence modeling with selective state spaces},
  author={Gu, Albert and Dao, Tri},
  booktitle={First conference on language modeling},
  year={2024}
}

@article{wang2024graph,
  title={Graph-mamba: Towards long-range graph sequence modeling with selective state spaces},
  author={Wang, Chloe and Tsepa, Oleksii and Ma, Jun and Wang, Bo},
  journal={arXiv preprint arXiv:2402.00789},
  year={2024}
}

@article{chen2025temporal,
  title={Temporal scene-object graph learning for object navigation},
  author={Chen, Lu and He, Zongtao and Wang, Liuyi and Liu, Chengju and Chen, Qijun},
  journal={IEEE Robotics and Automation Letters},
  year={2025},
  publisher={IEEE}
}

@article{meng2025context,
  title={Context-aware graph inference and generative adversarial imitation learning for object-goal navigation in unfamiliar environment},
  author={Meng, Yiyue and Guo, Chi and Li, Aolin and Luo, Yarong},
  journal={IEEE Robotics and Automation Letters},
  year={2025},
  publisher={IEEE}
}

@inproceedings{mnih2016asynchronous,
  title={Asynchronous methods for deep reinforcement learning},
  author={Mnih, Volodymyr and Badia, Adria Puigdomenech and Mirza, Mehdi and Graves, Alex and Lillicrap, Timothy and Harley, Tim and Silver, David and Kavukcuoglu, Koray},
  booktitle={International conference on machine learning},
  pages={1928--1937},
  year={2016},
  organization={PmLR}
}

@article{kolve2017ai2,
  title={Ai2-thor: An interactive 3d environment for visual ai},
  author={Kolve, Eric and Mottaghi, Roozbeh and Han, Winson and VanderBilt, Eli and Weihs, Luca and Herrasti, Alvaro and Deitke, Matt and Ehsani, Kiana and Gordon, Daniel and Zhu, Yuke and others},
  journal={arXiv preprint arXiv:1712.05474},
  year={2017}
}

@inproceedings{deitke2020robothor,
  title={Robothor: An open simulation-to-real embodied ai platform},
  author={Deitke, Matt and Han, Winson and Herrasti, Alvaro and Kembhavi, Aniruddha and Kolve, Eric and Mottaghi, Roozbeh and Salvador, Jordi and Schwenk, Dustin and VanderBilt, Eli and Wallingford, Matthew and others},
  booktitle={Proceedings of the IEEE/CVF conference on computer vision and pattern recognition},
  pages={3164--3174},
  year={2020}
}

@inproceedings{zhang2023layout,
  title={Layout-based causal inference for object navigation},
  author={Zhang, Sixian and Song, Xinhang and Li, Weijie and Bai, Yubing and Yu, Xinyao and Jiang, Shuqiang},
  booktitle={Proceedings of the IEEE/CVF Conference on Computer Vision and Pattern Recognition},
  pages={10792--10802},
  year={2023}
}

@inproceedings{xu2024aligning,
  title={Aligning knowledge graph with visual perception for object-goal navigation},
  author={Xu, Nuo and Wang, Wen and Yang, Rong and Qin, Mengjie and Lin, Zheyuan and Song, Wei and Zhang, Chunlong and Gu, Jason and Li, Chao},
  booktitle={2024 IEEE International Conference on Robotics and Automation (ICRA)},
  pages={5214--5220},
  year={2024},
  organization={IEEE}
}

@article{chi2025navigating,
  title={Navigating with Spatial Intelligence: A Survey of Scene Graph-Based Object Goal Navigation},
  author={Chi, GUO and Aolin, LI and Yiyue, MENG},
  journal={Wuhan University Journal of Natural Sciences},
  volume={30},
  number={5},
  pages={405--426},
  year={2025},
  publisher={Wuhan University}
}

@inproceedings{du2020learning,
  title={Learning object relation graph and tentative policy for visual navigation},
  author={Du, Heming and Yu, Xin and Zheng, Liang},
  booktitle={European Conference on Computer Vision},
  pages={19--34},
  year={2020},
  organization={Springer}
}

@inproceedings{zhang2021hierarchical,
  title={Hierarchical object-to-zone graph for object navigation},
  author={Zhang, Sixian and Song, Xinhang and Bai, Yubing and Li, Weijie and Chu, Yakui and Jiang, Shuqiang},
  booktitle={Proceedings of the IEEE/CVF international conference on computer vision},
  pages={15130--15140},
  year={2021}
}

@inproceedings{dang2022unbiased,
  title={Unbiased directed object attention graph for object navigation},
  author={Dang, Ronghao and Shi, Zhuofan and Wang, Liuyi and He, Zongtao and Liu, Chengju and Chen, Qijun},
  booktitle={Proceedings of the 30th ACM International Conference on Multimedia},
  pages={3617--3627},
  year={2022}
}

@article{sun2025enhancing,
  title={Enhancing multimodal-input object goal navigation by leveraging large language models for inferring room--object relationship knowledge},
  author={Sun, Leyuan and Kanezaki, Asako and Caron, Guillaume and Yoshiyasu, Yusuke},
  journal={Advanced Engineering Informatics},
  volume={65},
  pages={103135},
  year={2025},
  publisher={Elsevier}
}

@article{li2024stg,
  title={Stg-mamba: Spatial-temporal graph learning via selective state space model},
  author={Li, Lincan and Wang, Hanchen and Zhang, Wenjie and Coster, Adelle},
  journal={arXiv preprint arXiv:2403.12418},
  year={2024}
}

@article{sun2025memory,
  title={Memory-MambaNav: Enhancing object-goal navigation through integration of spatial--temporal scanning with state space models},
  author={Sun, Leyuan and Yoshiyasu, Yusuke},
  journal={Image and Vision Computing},
  volume={158},
  pages={105522},
  year={2025},
  publisher={Elsevier}
}

@article{liu2024robomamba,
  title={Robomamba: Efficient vision-language-action model for robotic reasoning and manipulation},
  author={Liu, Jiaming and Liu, Mengzhen and Wang, Zhenyu and An, Pengju and Li, Xiaoqi and Zhou, Kaichen and Yang, Senqiao and Zhang, Renrui and Guo, Yandong and Zhang, Shanghang},
  journal={Advances in Neural Information Processing Systems},
  volume={37},
  pages={40085--40110},
  year={2024}
}

@article{wei2022chain,
  title={Chain-of-thought prompting elicits reasoning in large language models},
  author={Wei, Jason and Wang, Xuezhi and Schuurmans, Dale and Bosma, Maarten and Xia, Fei and Chi, Ed and Le, Quoc V and Zhou, Denny and others},
  journal={Advances in neural information processing systems},
  volume={35},
  pages={24824--24837},
  year={2022}
}

@article{zhou2025mtil,
  title={Mtil: Encoding full history with mamba for temporal imitation learning},
  author={Zhou, Yulin and Lin, Yuankai and Peng, Fanzhe and Chen, Jiahui and Huang, Kaiji and Yang, Hua and Yin, Zhouping},
  journal={IEEE Robotics and Automation Letters},
  year={2025},
  publisher={IEEE}
}

@article{yang2018visual,
  title={Visual semantic navigation using scene priors},
  author={Yang, Wei and Wang, Xiaolong and Farhadi, Ali and Gupta, Abhinav and Mottaghi, Roozbeh},
  journal={arXiv preprint arXiv:1810.06543},
  year={2018}
}

@inproceedings{kwon2021visual,
  title={Visual graph memory with unsupervised representation for visual navigation},
  author={Kwon, Obin and Kim, Nuri and Choi, Yunho and Yoo, Hwiyeon and Park, Jeongho and Oh, Songhwai},
  booktitle={Proceedings of the IEEE/CVF international conference on computer vision},
  pages={15890--15899},
  year={2021}
}

@inproceedings{moghaddam2021optimistic,
  title={Optimistic agent: Accurate graph-based value estimation for more successful visual navigation},
  author={Moghaddam, Mahdi Kazemi and Wu, Qi and Abbasnejad, Ehsan and Shi, Javen},
  booktitle={Proceedings of the IEEE/CVF winter conference on applications of computer vision},
  pages={3733--3742},
  year={2021}
}

@inproceedings{zhu2017target,
  title={Target-driven visual navigation in indoor scenes using deep reinforcement learning},
  author={Zhu, Yuke and Mottaghi, Roozbeh and Kolve, Eric and Lim, Joseph J and Gupta, Abhinav and Fei-Fei, Li and Farhadi, Ali},
  booktitle={2017 IEEE international conference on robotics and automation (ICRA)},
  pages={3357--3364},
  year={2017},
  organization={ieee}
}

@inproceedings{wortsman2019learning,
  title={Learning to learn how to learn: Self-adaptive visual navigation using meta-learning},
  author={Wortsman, Mitchell and Ehsani, Kiana and Rastegari, Mohammad and Farhadi, Ali and Mottaghi, Roozbeh},
  booktitle={Proceedings of the IEEE/CVF conference on computer vision and pattern recognition},
  pages={6750--6759},
  year={2019}
}

@inproceedings{mayo2021visual,
  title={Visual navigation with spatial attention},
  author={Mayo, Bar and Hazan, Tamir and Tal, Ayellet},
  booktitle={Proceedings of the IEEE/CVF conference on computer vision and pattern recognition},
  pages={16898--16907},
  year={2021}
}

@article{du2021vtnet,
  title={Vtnet: Visual transformer network for object goal navigation},
  author={Du, Heming and Yu, Xin and Zheng, Liang},
  journal={arXiv preprint arXiv:2105.09447},
  year={2021}
}

@inproceedings{carion2020end,
  title={End-to-end object detection with transformers},
  author={Carion, Nicolas and Massa, Francisco and Synnaeve, Gabriel and Usunier, Nicolas and Kirillov, Alexander and Zagoruyko, Sergey},
  booktitle={European conference on computer vision},
  pages={213--229},
  year={2020},
  organization={Springer}
}

@inproceedings{he2016deep,
  title={Deep residual learning for image recognition},
  author={He, Kaiming and Zhang, Xiangyu and Ren, Shaoqing and Sun, Jian},
  booktitle={Proceedings of the IEEE conference on computer vision and pattern recognition},
  pages={770--778},
  year={2016}
}

@inproceedings{pennington2014glove,
  title={Glove: Global vectors for word representation},
  author={Pennington, Jeffrey and Socher, Richard and Manning, Christopher D},
  booktitle={Proceedings of the 2014 conference on empirical methods in natural language processing (EMNLP)},
  pages={1532--1543},
  year={2014}
}

\end{document}